\documentclass[10pt,twocolumn,letterpaper]{article}

\usepackage{iccv} 

\usepackage{pifont}

\usepackage[inline]{enumitem}
\definecolor{iccvblue}{rgb}{0.21,0.49,0.74}
\usepackage[pagebackref,breaklinks,colorlinks,allcolors=iccvblue]{hyperref}

\def\paperID{12829} 
\def\confName{ICCV}
\def\confYear{2025}

\usepackage[utf8]{inputenc} 
\usepackage[T1]{fontenc}    
\usepackage{hyperref}       
\usepackage{url}            
\usepackage{booktabs}       
\usepackage{amsfonts}       
\usepackage{nicefrac}       
\usepackage{floatrow}
\usepackage{microtype}      

\usepackage{times}
\usepackage{epsfig}
\usepackage{subcaption}
\usepackage{graphicx}
\usepackage{booktabs}

\usepackage{multirow}
\usepackage[normalem]{ulem}
\usepackage{xcolor}
\usepackage{colortbl}
\definecolor{Gray}{gray}{0.93}
\usepackage[linesnumbered,ruled]{algorithm2e}
\usepackage{amsmath}
\usepackage[capitalize]{cleveref}
\usepackage{booktabs}
\usepackage{varwidth}
\usepackage{paralist}
\usepackage{enumerate}
\usepackage{enumitem}
\usepackage{amssymb}
\usepackage{pifont}
\usepackage{bm}
\usepackage{amsmath}
\usepackage{multirow}
\usepackage{caption}
\usepackage{wrapfig}
\usepackage[export]{adjustbox}
\usepackage{subfiles}

\newlength\MyHangIndent
\usepackage{xcolor}    

\usepackage{enumitem}   
\usepackage{hyperref}   
\usepackage{graphicx}   
\usepackage{xcolor}     
\usepackage{booktabs}   

\crefname{equation}{Eq.}{Eq.}
\crefname{section}{Section}{Sections}
\crefname{subsection}{Section}{Sections}
\crefname{subsubsection}{Section}{Sections}
\crefname{figure}{Figure}{Figures}
\crefname{table}{Table}{Tables}
\crefname{subfigure}{Figure}{Figures}
\crefname{algocf}{Algorithm}{Algorithms}
\useunder{\uline}{\ul}{}

\definecolor{customred}{RGB}{200,42,42}
\definecolor{sfblue}{HTML}{0d9dda}

\title{BLIP-3: A Family of Open Large Multimodal Models}

\author{
\baselineskip=3pt
\small
Le Xue$^{1}$$^{\circ}$ \hspace*{0.5em}
Manli Shu$^{1}$$^{\circ}$ \hspace*{0.5em}
Anas Awadalla$^{1,3}$$^{*}$ \hspace*{0.5em}
Jun Wang$^{1}$$^{*}$ \hspace*{0.5em}
An Yan$^{1}$$^{*}$ \hspace*{0.5em}
Senthil Purushwalkam$^{1}$$^{*}$ \hspace*{0.5em} \vspace{-3pt}\\
\small
Honglu Zhou$^{1}$$^{*}$ \hspace*{0.5em}
Viraj Prabhu$^{1}$$^{*}$ \hspace*{0.5em}
Yutong Dai$^{1}$$^{*}$ \hspace*{0.5em}
Michael S Ryoo$^{1}$$^{*}$ \hspace*{0.5em}
Shrikant Kendre$^{1}$$^{*}$ \hspace*{0.5em}
Jieyu Zhang$^{1,3}$$^{*}$ \hspace*{0.5em} \vspace{-3pt}\\
\small
Shaoyen Tseng$^{2}$$^{*}$ \hspace*{0.5em}
Gustavo A Lujan-Moreno$^{2}$$^{*}$ \hspace*{0.5em}
Matthew L Olson$^{2}$$^{*}$ \hspace*{0.5em}
Musashi Hinck$^{2}$$^{*}$ \hspace*{0.5em}
David Cobbley$^{2}$$^{*}$ \hspace*{0.5em}
Vasudev Lal$^{2}$$^{*}$ \hspace*{0.5em} \vspace{-3pt}\\
\small
Can Qin$^{1}$ \hspace*{0.5em}
Shu Zhang$^{1}$ \hspace*{0.5em}
Chia-Chih Chen$^{1}$ \hspace*{0.5em}
Ning Yu$^{1}$ \hspace*{0.5em}
Juntao Tan$^{1}$ \hspace*{0.5em}
Tulika Manoj Awalgaonkar$^{1}$ \hspace*{0.5em} \vspace{-3pt}\\
\small
Shelby Heinecke$^{1}$$^{\dagger}$ \hspace*{0.5em}
Huan Wang$^{1}$$^{\dagger}$ \hspace*{0.5em}
Yejin Choi$^{3}$$^{\dagger}$ \hspace*{0.5em}
Ludwig Schmidt$^{3}$$^{\dagger}$ \hspace*{0.5em} \vspace{-3pt}\\
\small
Zeyuan Chen$^{1}$$^{\dagger}$ \hspace*{0.5em}
Silvio Savarese$^{1}$$^{\dagger}$ \hspace*{0.5em}
Juan Carlos Niebles$^{1}$$^{\dagger}$ \hspace*{0.5em}
Caiming Xiong$^{1}$$^{\dagger}$ \hspace*{0.5em}
Ran Xu$^{1}$$^{\dagger}$ \vspace{3pt} \\
\small
$^{1}$Salesforce AI Research \hspace*{0.1em} $^{2}$Intel Labs\hspace*{0.1em} $^{3}$University of Washington\hspace*{0.1em}
\vspace{-3pt} \\
\small
\texttt{\{lxue, ssavarese, jniebles, cxiong, ran.xu\}@salesforce.com} \vspace{-3pt} \\
\small
{$^{\circ}$}First Authors; $^{*}$Core Authors; $^{\dagger}$Senior Authors \\
\small
\href{https://www.salesforceairesearch.com/opensource/xGen-MM/index.html}{\textcolor{sfblue}{Project Page}}
}
\begin{document}

\maketitle

\begin{abstract}
\looseness -1 This paper introduces BLIP-3, an open framework for developing Large Multimodal Models (LMMs). The framework comprises meticulously curated datasets, a training recipe, model architectures, and a resulting suite of LMMs. We release 4B and 14B models, including both the pre-trained base model and the instruction fine-tuned ones. Our models undergo rigorous evaluation across a range of tasks, including both single and multi-image benchmarks.
Our models demonstrate competitive performance among open-source LMMs with similar model sizes.
Our resulting LMMs demonstrate competitive performance among open-source LMMs with similar model sizes, with the ability to comprehend interleaved image-text inputs.
Our training code, models, and all datasets used in this work, including the three large-scale datasets we create and the preprocessed ones, will be open-sourced to better support the research community.
\begin{figure*}[h!]
    \centering
    \includegraphics[width=1\textwidth]{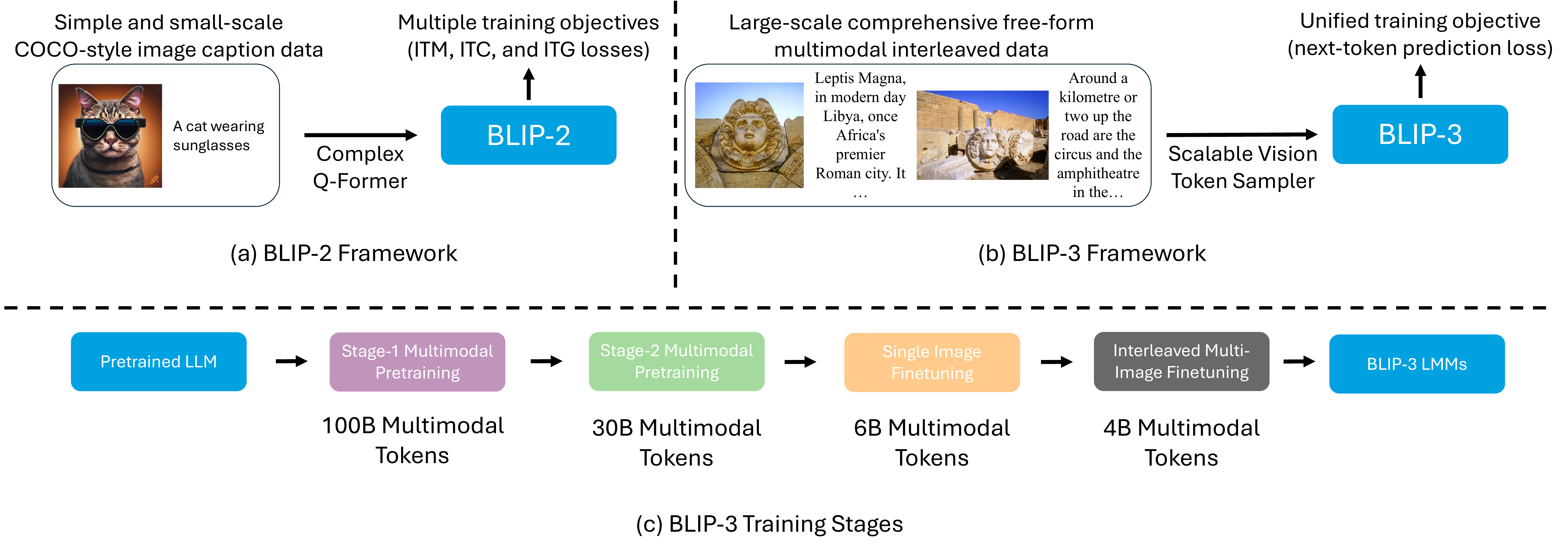}
    \caption{\small{
    \textbf{We introduce BLIP-3, a framework (b) for developing Large Multimodal Models (LMMs).} Our framework improves upon BLIP-2 (a) \cite{blip2} by (1) increasing the richness, scale, and diversity of training data, (2) replacing the Q-Former layers with a more scalable vision token sampler, and (3) simplifying the training process via the unification of the training objectives to a single loss at every training stage. (c) illustrates the stage-by-stage training process, from initializing with a pre-trained LLM to the final LMMs.
    }}
    \label{fig:teaser}
\vspace{-3mm}
\end{figure*}
\end{abstract}

\section{Introduction}



\looseness -1 Large Multimodal Models (LMMs) have attracted significant attention with their potential applications and emergent capabilities. Recent advancements in both proprietary models~\cite{gpt4v, Gemini, gpt4o, mm1.5} and open-source LMMs~\cite{blip,blip2,instructblip,liu2023llava,mckinzie_mm1_2024, vila, idefics2, minicpm, vila1.5, pixtral, llama3, llama3p1,cambrian, xinstructblip} highlight the rapid progress and growing interest in this field.
However, despite these advancements, a significant gap remains between open-source and proprietary models, particularly in terms of accessibility to good base models, training recipes, and curated datasets. 
While there are many strong open-weight models~\cite{abdin_phi3_2024, pixtral, llama3}, not many of them disclose their training details or release their full training data recipe (including training code).
This lack of transparency limits the broader research community’s ability to replicate, analyze, and improve LMMs. Additionally, many of these models do not contribute new datasets, further restricting progress. These constraints hinder transparency and innovation, preventing open-source communities from fully leveraging and advancing LMM research.

Recent works have demonstrated that large-scale and high-quality data are essential for training robust LMMs~\cite{liu2023llava, mckinzie_mm1_2024, vila,idefics2, mint1t, kale}. BLIP-2~\cite{blip2} was one of the pioneering efforts in exploring LMMs, which leveraged synthetic data to achieve impressive results at the time (Figure \ref{fig:teaser} (a)). However, the data used in BLIP-2 lacks the scale, quality, and diversity required to reach competitive performance compared to more modern LMMs nowadays. In addition, BLIP-2 employs an intricate Q-Former~\cite{blip2} architecture to bridge the vision and language modalities, coupled with a suite of complex training objectives (ITM, ITC, and ITG losses), both of which pose obstacles for larger-scale training. Moreover, BLIP-2 supports only single-image input, whereas interleaved multimodal data formats are the most natural form of multimodal data~\cite{laurenccon2024obelics}.

\begin{figure}[t]
    \centering
    \includegraphics[width=1.0\linewidth]{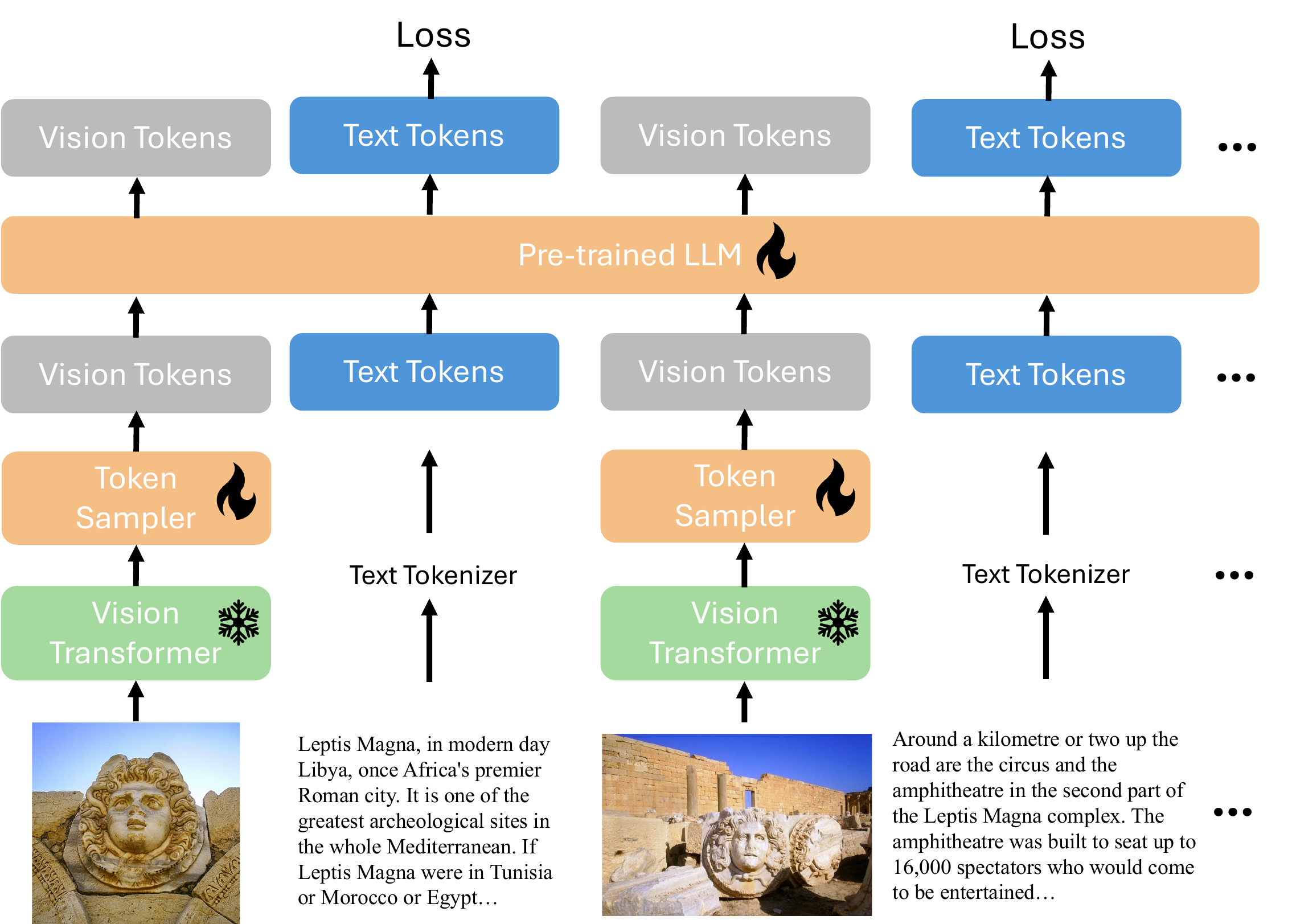}
    \caption{\small{\textbf{Overview of the BLIP-3 architecture.} Free-form interleaved images and texts from the ensembled interleaved and caption datasets are input into the framework, with each modality undergoing a separate tokenization process to be fed into the pre-trained LLM in natural order. A standard auto-regressive loss is then applied to the text tokens. The Vision Transformer is kept frozen during training, while all other parameters, including the vision token sampler and the pre-trained LLM, are trained.}}
    \label{fig:overview}
\end{figure}

In response to these challenges, we introduce BLIP-3 (Figure \ref{fig:teaser} (b)), a new framework designed to scale up LMM training by utilizing an ensemble of multimodal interleaved datasets, curated caption datasets, and other publicly available datasets~\cite{cc12m,vg,sbu,datacomp, idl}. In BLIP-3, as illustrated in Figure \ref{fig:overview}, we streamline the model architecture by replacing the Q-Former~\cite{blip2} with a more scalable vision token sampler~\cite{flamingo} and simplifying the training objectives to focus solely on the auto-regressive loss of text tokens in a multimodal context. Our primary focus is on dataset curation and scaling up the training data.
Recently, several high-quality, large-scale datasets have been introduced, such as MINT-1T\cite{mint1t}, a trillion-token-scale interleaved dataset, and BLIP3-KALE\cite{kale}, a knowledge-augmented dataset with high-quality dense captions. In this paper, we introduce three additional specialized datasets: BLIP3-OCR-200M, a large-scale dataset with dense OCR annotations for general base-resolution OCR understanding; BLIP3-GROUNDING-50M, a large-scale visual grounding dataset; and BLIP3-OCR-HD-30M, a high-resolution dataset with high-quality OCR annotations, designed for high-resolution pre-training.

\looseness -1 In addition to these datasets, we are committed to open-sourcing the series of models developed in this work, including both the pre-trained base models and instruction-tuned models. Along with the model release, we also provide our training code. By making these resources publicly available, we aim to make LMM research and development more accessible to the community, and we encourage researchers and practitioners to use our models and datasets to understand and further explore the potential and emergent capabilities of LMMs. 


\section{Related Work}
\label{sec:related_work}

\looseness -1 Recent advancements in Large Multimodal Models (LMMs) have explored different architectural designs. We roughly divide them into two main categories: the cross-attention style~\cite{flamingo,openflamingo} and the self-attention style. The cross-attention approach, exemplified by models like Flamingo~\cite{flamingo, openflamingo} and Llama 3.1~\cite{llama3p1}, integrates vision and language modalities by injecting visual information into layers inside LLMs using a cross-modal cross-attention module.
The self-attention approach~\cite{blip2,instructblip,liu2023llava,qwen-vl,palix,palm-e,kosmos-1,kosmos-2,liu2023improvedllava,mplug-owl,mplug-owl2,multimodal-gpt,honeybee,sphinx,emu2}, on the other hand, offers a more streamlined solution. This approach connects pre-trained language models to visual inputs using lightweight connectors. Both visual and textual inputs are tokenized and concatenated in the same input sequence to LLM, and the LLM learns to align both modalities through self-attention. The self-attention design has become prevalent recently, which has been shown to be effective on a wide range of LMMs~\cite{liu2024llavanext, abdin_phi3_2024, pixtral, yao2024minicpmvgpt4vlevelmllm, qwen-vl} . For BLIP-3, we also adopt the self-attention design due to its simplicity.

\looseness -1 Training methodologies for LMMs typically follow one of the two strategies. The first one uses a light pre-training procedure and heavily relies on visual instruction tuning, as seen in the LLaVA series~\cite{liu2023llava, liu2023improvedllava}. Extensive research has been conducted on creating effective instruction-tuning data for a variety of tasks~\cite{multimodal-gpt,sharegpt4v,m3it,svit,lvis-instruct4v,mimic-it}. The second strategy involves extensive pre-training on large-scale, diverse datasets, followed by visual instruction fine-tuning. This approach, used by models such as MM1 series~\cite{mckinzie_mm1_2024, mm1.5} and Idefics2~\cite{idefics2}, infuses a broad knowledge into the model during the pre-training stage, and then fine-tune it for better instruction following ability and better alignment. While MM1 series~\cite{mckinzie_mm1_2024, mm1.5} provides extensive ablations and studies on the recipes aimed at improving LMMs, it releases limited resources for practitioners to reproduce the model (their models and datasets are not open-sourced). 


\looseness -1 In this work, we introduce BLIP-3, an open LMM family encompassing a series of models, data recipes, training code, and three new large-scale foundational multimodal datasets. Unlike previous works, BLIP-3 is designed to enable and advance future research in this area by providing comprehensive resources for the community.


\section{Model Architecture}
\label{sec:modeling}
\paragraph{Architecture Overview.}
\looseness -1 As illustrated in Figure \ref{fig:overview}, the BLIP-3 framework adopts an architecture consisting of a ViT \cite{dfn, siglip}, a vision token sampler (perceiver resampler~\cite{flamingo}) to downsample the image embeddings, and a pre-trained LLM (Phi-3 series~\cite{abdin_phi3_2024}). The input to the model can be free-form multimodal interleaved texts and vision tokens from the diverse multimodal data sources we ensemble. 

\vspace{5pt}
\noindent\textbf{Any-Resolution Vision Token Sampling.}
As proved effective in recent LMMs~\cite{liu2024llavanext, zhang_ferret-v2_2024,dong_internlm-xcomposer2-4khd_2024}, we adopt a dynamic high-resolution (\textit{i.e.}, ``any-resolution'') image encoding strategy at both the Stage-2 pre-training and the fine-tuning stages. We enable higher-resolution image understanding with image patch-wise encoding. The patch-wise encoding preserves the resolution of the original image as much as possible by splitting a single image into multiple patches and encoding them separately. Following the previous convention, we concatenate the encoded image patches with a downsized original image that provides global information.

\looseness -1 In the VL connector, we use a perceiver resampler to downsample the vision tokens. With any-resolution image encoding, we perform the downsampling for each image patch (including the downsized original image) independently. The downsampler vision tokens are then concatenated together and sent to the LLM. With the downsampling in our VL connector, we can reduce the sequence length of vision tokens by a factor of five or more depending on the number of query tokens in the perceiver resampler. We provide ablation studies on different token sampling strategies in Section~\ref{sec:sft-ablation-studies}.

\section{Training}
\vspace{5pt}
\noindent\textbf{Stage-1 Base Resolution Pre-training.}
The first stage of pre-training focuses on base resolution image-text alignment using a diverse dataset mixture comprising open-source datasets and two newly created datasets from BLIP-3 -- BLIP3-OCR-200M and BLIP3-GROUNDING-50M, all of which will be made publicly available. In this stage, the base model is trained on approximately 100 billion multimodal tokens from the ensembled dataset. To ensure computational efficiency, the pre-training resolution is set to 384×384 pixels, aligning with the default resolution of SigLIP~\cite{siglip}.

\vspace{5pt}
\noindent\textbf{Stage-2 High Resolution Pre-training.}
Building upon the pre-trained checkpoint from Stage 1, the second stage incorporates additional high-quality, high-resolution datasets, including one newly created dataset from BLIP-3 -- BLIP3-OCR-HD-30M, to further enhance the model’s capability. To maintain consistency with the supervised fine-tuning (SFT) stage, we adopt the same any-resolution vision token sampling strategy introduced in section~\ref{sec:modeling}. We set the any-resolution grid to support up to 12 image patches with a maximum side length of 1536 pixels on the longer side. Further details and ablation studies are provided in section~\ref{sec:stage-2-pretraining-ablation-studies}.

\vspace{5pt}
\noindent\textbf{Single-Image Supervised Fine-tuning.}
Then, we fine-tune the pre-trained model on a collection of publicly available instruction-following datasets~\cite{idefics2,liu2023improvedllava, cambrian}. During fine-tuning, we employ the same any-resolution vision token sampling strategy as in Stage 2, enabling the model to process high-resolution images effectively, including text-rich document-style data. The following sections provide additional technical details on the fine-tuning procedure.

\vspace{5pt}
\noindent\textbf{Interleaved Multi-Image Supervised Fine-tuning.}
We conduct a second-stage fine-tuning on the instruction fine-tuned model on a mixture of multi-image and single-image instruction-following samples. The goal of this second-stage fine-tuning is to enhance the model's ability to comprehend interleaved image-text input, which is helpful for multimodal in-context learning, multi-image question answering, and many more practical use cases. For the multi-image fine-tuning, we also adopt the any-resolution vision token sampling strategy, the same as in the previous SFT stage.



\section{Data}
\label{sec:data}
\subsection{Pre-training Data Recipe}
\begin{figure*}[h]
    \centering
    \includegraphics[width=0.9\linewidth]{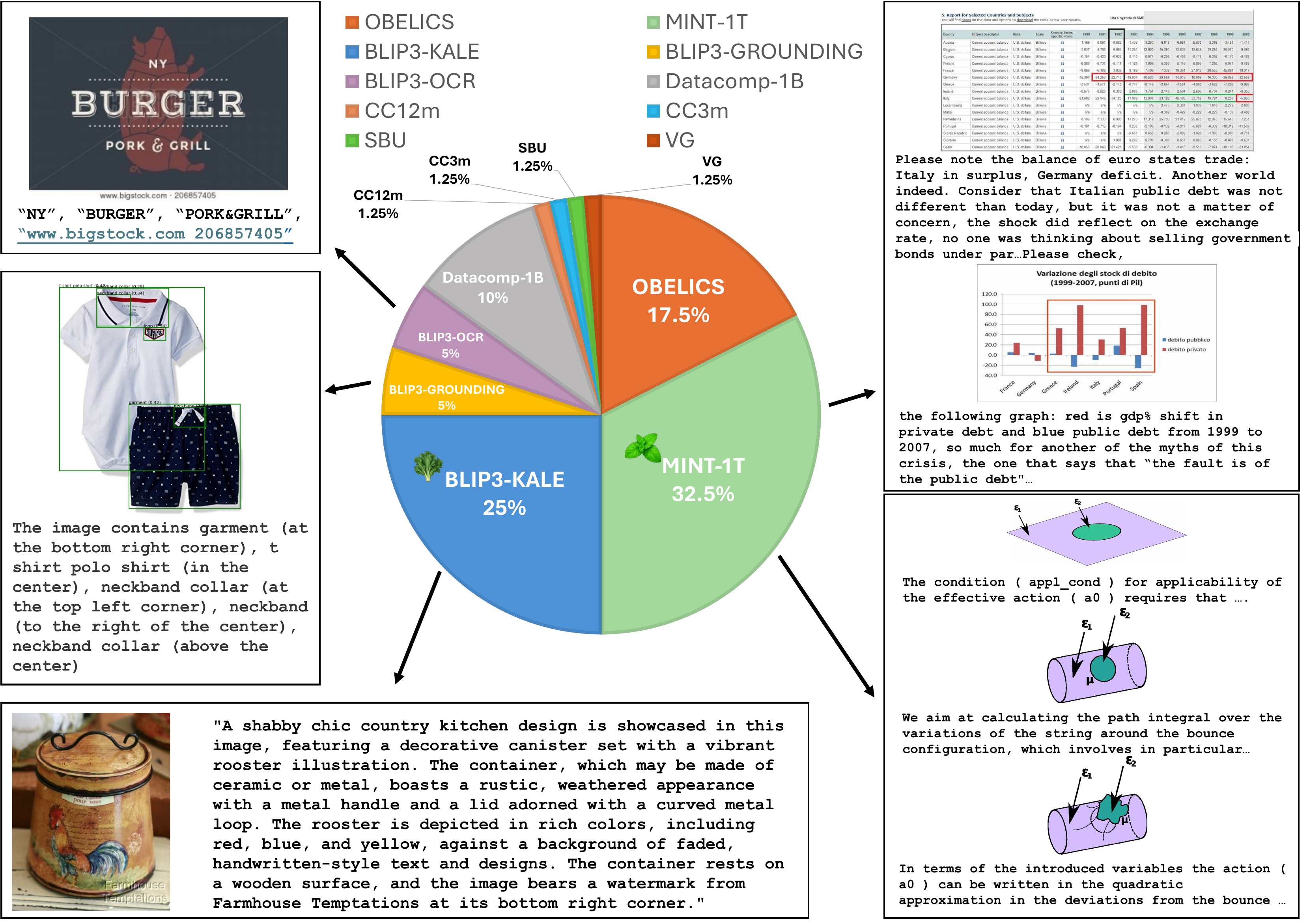}
    \caption{\small{\textbf{Overview of BLIP-3 Stage-1 pre-training datasets base recipe.}}}
    \label{fig:pretrain_datasets}
\end{figure*}

Our Stage-1 pre-training recipe is detailed in Figure \ref{fig:pretrain_datasets}. In BLIP-3, we pre-train on a diverse set of multimodal datasets using the specified sampling ratios.

For Stage-2 pre-training, we use the Stage-1 data recipe as a base while incorporating specialized high-resolution, text-rich datasets. Specifically, we replace BLIP3-OCR-200M with BLIP3-OCR-HD-30M and introduce the IDL~\cite{idl} dataset, each making up 10\% of the total training data. To preserve the overall structure of the dataset mixture, we adjust the sampling ratios of the remaining caption datasets by reducing BLIP3-KALE from 25\% to 10\%, Datacomp-1B from 10\% to 2.5\%, and BLIP3-GROUNDING-50M from 5\% to 2.5\%, while keeping the rest unchanged.

\noindent\textbf{Interleaved Dataset Mixture.}
We combine MINT-1T (including its HTML, PDF, and ArXiv subsets) with OBELICS (HTML only) to create a more diverse and comprehensive dataset mixture that covers a broader range of domains.
\vspace{-5mm}
\paragraph{1. MINT-1T~\cite{mint1t}.}{A trillion-token-scale multimodal interleaved dataset, containing data sources from HTML, PDF, and ArXiv. As evidenced by MM1~\cite{mckinzie_mm1_2024} and Idefics2~\cite{idefics2}, such multimodal interleaved datasets are essential for scaling up large multimodal model training and enabling fundamental capabilities like multimodal in-context learning. In BLIP-3 we mix its HTML, PDF, and ArXiv subsets in a 7:5:1 ratio.}
\vspace{-5mm}
\paragraph{2. OBELICS~\cite{idefics2}.}{A multimodal interleaved dataset constructed solely from HTML documents. It differs slightly in domain coverage from MINT-1T due to the specific preprocessing steps adopted.}



\vspace{10pt}
\noindent\textbf{Caption Dataset Mixture.}
We integrate a diverse range of caption datasets, with the following details.
\begin{figure*}[!t]
    \centering
    \includegraphics[width=0.95\linewidth]{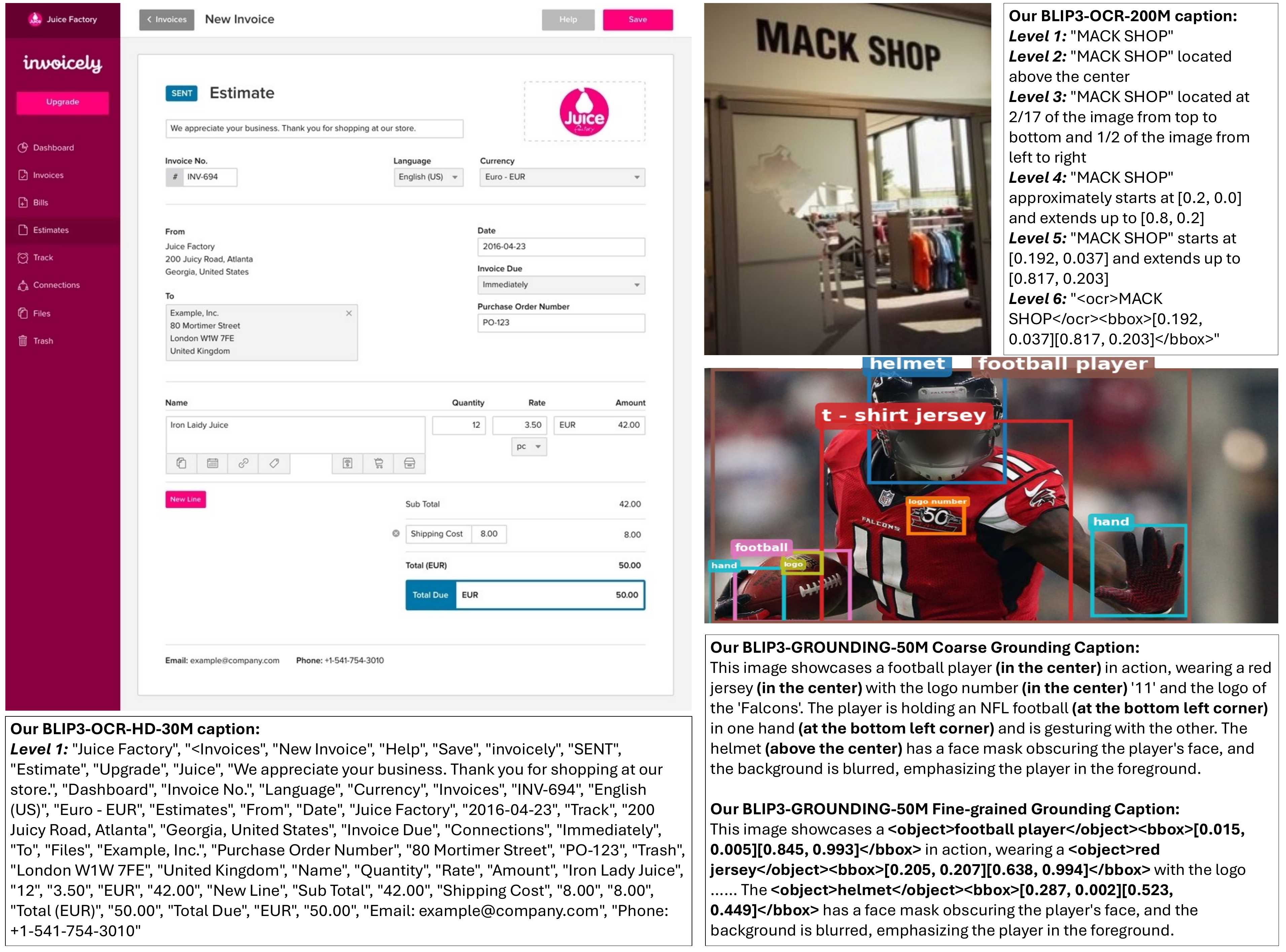}
    \caption{\small{\textbf{Samples from BLIP3-OCR-200M.} We extract six levels of OCR granularity, with and without bounding boxes. OCR-related captions in BLIP-3 are preprocessed to remove filler phrases like 'the text' within the dataloader, which we find improving OCR benchmarks' performance. 
    \small{\textbf{Samples from BLIP3-OCR-HD-30M.}} Structured similarly to BLIP3-OCR-200M, this dataset contains high-resolution images annotated with high-resolution OCR tools for enhanced high resolution OCR understanding. Only Level 1 is shown for simplicity.
    \small{\textbf{Samples from BLIP3-GROUNDING-50M.}} A large-scale dataset of images and corresponding captions containing localization information about objects. Furthermore, we release the associated object bounding box data to facilitate the creation of captions with custom templates.}}

\label{fig:blip3_ocr}
\end{figure*}

\vspace{-5mm}
\paragraph{1. \textbf{BLIP3-KALE~\cite{kale}}.} An open-source large-scale curated high-quality caption dataset, which we use as a base for general detailed caption datasets.
\vspace{-5mm}
\paragraph{2. \noindent\textbf{BLIP3-OCR-200M}.} A curated large-scale OCR dataset with 200 million images from Datacomp-
1B\cite{datacomp}. For each image, we use the off-the-shelf OCR engine~\cite{paddleocr} for OCR annotation. Text segments in a caption like \textit{"... text ..."} are modified to include OCR information as \textit{"... text ( \texttt{ocr\_info} ) ..."}, where \texttt{ocr\_info} contains normalized bounding box coordinates for the extracted text, specifying its exact position within the image in the format "\texttt{<bbox>}$x_1, y_1, x_2, y_2$\texttt{</bbox>}". We have multiple granularities of OCR information, including with and without bounding box data. In BLIP-3 pre-training, we only utilize textual information without bounding box data. 
\vspace{-5mm}
\paragraph{3. \noindent\textbf{BLIP3-GROUNDING-50M}.} A curated large-scale grounding dataset to enhance the ability to ground semantic concepts in visual features, which is crucial for tasks like understanding referring expressions~\cite{yu2016modeling} (e.g., "the object to the left of the dog"). We curate a dataset of 50 million images from Datacomp-1B~\cite{datacomp}. For each image, we identify objects and their location information using one of the state-of-the-art open-world image tagging~\cite{zhang2024recognize} and object detection models~\cite{liu2023grounding}. Objects mentioned in a caption like \textit{"... object ..."} are modified to include grounding information as \textit{" ... object ( \texttt{grounding\_info} ) ..."}, where \texttt{grounding\_info} contains bounding box information in one of three formats, each capturing a different granularity of localization: (1) \texttt{<bbox>$x_1, y_1, x_2, y_2$</bbox>}, (2) "starts at $(x_1, y_1)$ and extends up to $(x_2, y_2)$", or (3) "top-left corner of the image".

\vspace{-5mm}
\paragraph{4. \noindent\textbf{BLIP3-OCR-HD-30M}.} A high-resolution OCR dataset curated for Stage-2 pre-training. Different from BLIP3-OCR-200M, which focuses on more general OCR data with fewer resolution constraints, this dataset is filtered to only contain images of which both the height and width are larger than 512 pixels. On the selected high-resolution images, we run PaddleOCR with a high-resolution setup to obtain more accurate OCR annotations. Our empirical study shows that running the OCR annotation process with a high-res setting can help get a more complete and accurate set of OCR annotations that matches its resolution, which is more useful for the Stage-2 pre-training. This process yields a total of 30 million high-resolution samples with detailed OCR annotations that match the image quality. 
\vspace{-5mm}
\paragraph{5.\textbf{Other Datasets Mixture.}} We also include other publicly available datasets such as uncurated Datacomp-1B~\cite{datacomp} image-text pairs, CC12M~\cite{cc12m}, CC3M~\cite{cc12m}, VG~\cite{vg}, SBU~\cite{sbu}, and IDL~\cite{idl}.

    
\subsection{Supervised Fine-tuning Data Recipe} 
The datasets used in the fine-tuning stage are from public datasets of different domains~\cite{liu2023improvedllava, idefics2, cambrian,yan2024list, sharegpt4v,textvqa,OCR-VQA,A-OKVQA,GQA,DocVQA,ChartQA,DVQA,AI2D,CLEVR,ScienceQA, provision}. 
In addition to the multi-modal image-text data, we also mix in pure text instruction-following data ~\cite{OpenOrca, Orca-Math} during visual instruction tuning.
Ultimately, we collect a mixture of 3 million publicly available instruction-tuning samples, on which we fine-tune our model for one epoch.

The multi-image instruction tuning stage starts with a model fine-tuned on single-image data. We use a mixture of public multi-image / interleaved image-text instruction data~\cite{jiang_mantis_2024, liu2024mmdumultiturnmultiimagedialog}. To prevent the model from deteriorating on single-image capabilities, we reuse a subset of single-image datasets used in the previous fine-tuning stage and mix them into the multi-image training data. 

\section{Experiments}

\begin{table*}[!h]
    \vspace{5pt}
      \centering
      \resizebox{0.95\linewidth}{!}{
    \begin{tabular}{@{}lcccclllllllllllll@{}}
    \toprule
    \multirow{2}{*}{Model} 
    & \multicolumn{4}{c}{Accessibility} 
    & \rotatebox[origin=c]{90}{SEED-IMG} 
    & \rotatebox[origin=c]{90}{MMB(dev)} 
    & \rotatebox[origin=c]{90}{MMStar} 
    & \rotatebox[origin=c]{90}{MME(norm)} 
    & \rotatebox[origin=c]{90}{RWQA} 
    & \rotatebox[origin=c]{90}{MMVet} 
    & \rotatebox[origin=c]{90}{MMMU(val)} 
    & \rotatebox[origin=c]{90}{MathVista} 
    & \rotatebox[origin=c]{90}{TextVQA} 
    & \rotatebox[origin=c]{90}{OCRBench} 
    & \rotatebox[origin=c]{90}{POPE} 
    & \rotatebox[origin=c]{90}{HalBench} 
    & \cellcolor[HTML]{E7FAFE}\rotatebox[origin=c]{90}{Average} \\ 
    \cmidrule(lr){2-5}
    & \begin{tabular}[l]{@{}c@{}}open-\\weight\end{tabular} & \begin{tabular}[l]{@{}c@{}}open-\\data\end{tabular} & \begin{tabular}[l]{@{}c@{}}open-\\code \end{tabular}  & \begin{tabular}[l]{@{}c@{}}data\\contribution \end{tabular}
    & & & & & & & & & & & & & \\
    & & & & & & & & & & & & & & & &\\\midrule
    \color{gray}GPT-4V~\cite{gpt4v}  &    \color{gray}   $\times$   &   \color{gray}    $\times$  &   \color{gray}   $\times$   &   \color{gray}   $\times$  & \color{gray} 72.0 & \color{gray} 80.8  & \color{gray} 49.7   & \color{gray} 63.3   & \color{gray} 61.4   & \color{gray} 49.0  & \color{gray} 53.8   & \color{gray} 48.2   & \color{gray} 78.0  & \color{gray} 67.8   & \color{gray} 81.8  & \color{gray} 39.3  &\cellcolor[HTML]{E7FAFE} \color{gray} 62.1     \\\midrule
    \textit{3B-5B Models}    &   &  & & &  &    &   & & &  &  &    &  &   &   &  &\\\midrule
    MM1.5-3B~\cite{mm1.5}              &       $\times$       &       $\times$     &      $\times$   &      $\times$      & 72.4             & -             & -          & 64.2         & 56.9          & 41.0        & 37.1        & 44.4           & 76.5           & 65.7      & 88.1        & -          &\cellcolor[HTML]{E7FAFE} -        \\
    Phi-3-vision~\cite{abdin_phi3_2024}             &       \checkmark     &       $\times$     &      $\times$   &      $\times$      & 71.0             & 73.6          & 47.9       & 55.3         & 58.8          & 43.2        & 46.1        & 45.1           & 73.3           & 63.7      & 83.5        & 39.0       &\cellcolor[HTML]{E7FAFE} 58.4     \\
    Phi-3.5-vision~\cite{abdin_phi3_2024}            &       \checkmark     &       $\times$     &      $\times$   &      $\times$      & 69.6             & 81.9          & 47.3       & 65.6         & 53.5          & 43.2        & 43.0        & 43.9           & 72.0           & 59.9      & 86.1        & 39.6       &\cellcolor[HTML]{E7FAFE} 58.8     \\
    MiniCPM-V-2.0~\cite{yao2024minicpmvgpt4vlevelmllm}            &       \checkmark     &       $\times$     &      $\times$   &      $\times$      & 67.1             & 69.1          & 39.1       & 64.6         & 55.8          & 41.0        & 38.2        & 39.8           & 73.2           & 59.6      & 86.3        & 36.1       &\cellcolor[HTML]{E7FAFE} 55.8     \\
    VILA-1.5-3B~\cite{vila}                               &       \checkmark     &       \checkmark   &      \checkmark &    $\times$        & 67.9             & 62.4          & 40.3       & 58.5         & 53.3          & 38.5        & 34.1        & 30.6           & 58.1           & 43.7      & 86.9        & 31.2       &\cellcolor[HTML]{E7FAFE} 50.5     \\
    \rowcolor[HTML]{E7FAFE}      
    BLIP-3-4B (ours)                           &       \checkmark    &        \checkmark   &      \checkmark &      \checkmark    & 71.9             & 74.7          & 45.3       & 62.5         & 61.0          & 41.2        & 42.3        & 44.5           & 76.5           & 66.0      & 87.9        & 36.5       &\cellcolor[HTML]{E7FAFE} \textbf{59.2}     \\\midrule
    \textit{10B-15B Models}    &   & &  &  &  &    &   & & &  &  &    &  &   &   &  &\\\midrule      
    Pixtral-12B~\cite{pixtral}         &       \checkmark     &       $\times$     &     $\times$    &     $\times$       & 71.5             & 77.9          & 54.5       & 68.6         & 65.4          & 58.5        & 51.1        & 56.3           & 75.7           & 72.2      & 84.2        & 47.0       &\cellcolor[HTML]{E7FAFE} \textbf{64.9}     \\
    Llama-3.2-11B-V-inst.~\cite{llama3p1}        &       \checkmark     &       $\times$     &     $\times$    &     $\times$       & 72.7             & 68.0          & 49.8       & 65.0         & 63.3          & 57.6        & 48.0        & 47.7           & 54.1           & 75.3      & 88.1        & 40.3       &\cellcolor[HTML]{E7FAFE} 60.8     \\
    VILA-1.5-13B~\cite{vila}             &       \checkmark     &       \checkmark   &     \checkmark  &     $\times$       & 68.0             & 74.4          & 44.2       & 61.4         & 53.3          & 45.0        & 41.1        & 42.3           & 61.1           & 45.7      & 85.0        & 39.3       &\cellcolor[HTML]{E7FAFE} 55.1     \\
    Cambrian-13B~\cite{cambrian}       &       \checkmark     &       \checkmark   &     \checkmark  &     \checkmark     & 73.2             & 73.2          & 47.1       & 67.0         & 58.6          & 48.9        & 41.6        & 47.7           & 72.8           & 61.0      & 86.8        & 39.4       &\cellcolor[HTML]{E7FAFE} 59.8     \\
    \rowcolor[HTML]{E7FAFE}      
    BLIP-3-14B (ours)                          &       \checkmark     &       \checkmark   &     \checkmark  &     \checkmark     & 73.4    & 77.5          & 48.7       & 68.1         & 64.4          & 44.5        & 47.0        & 45.4           & 79.0           & 71.1      & 86.0        & 38.7       &\cellcolor[HTML]{E7FAFE} 62.0     \\ \bottomrule
    \end{tabular}
    \caption{\label{tab:sft-single-all}\textbf{Evaluation on single-image benchmarks.} For benchmarks where a open-source model doesn't report its official score, we use the score reported in a third-party leaderboard\protect\footnotemark or run it ourselves using the evaluation codebase~\cite{vlmevalkit} for a fair comparison. We also include the GPT-4V ({\small\texttt{gpt-4-1106-preview}}) performance (provided by the evaluation codebase) as a reference in the first row.}}
    \end{table*}

    \footnotetext{\url{https://huggingface.co/spaces/opencompass/open_vlm_leaderboard}}
\label{sec: sft-main-results}
We evaluate our models (4B and 14B) on a comprehensive suite of multimodal benchmarks, assessing the model's ability from multiple perspectives. Our evaluation covers general VQA benchmarks~\cite{li2023seed,liu2023mmbench,mme,mmstar,realworldQA}, domain knowledge~\cite{mmmu, mathvista}, OCR ability~\cite{textvqa,ocrbench}, and hallucination~\cite{guan2024hallusionbench, li2023evaluating}. 
For models fine-tuned on interleaved multi-image datasets, we also evaluate their performance on common multi-image benchmarks~\cite{jiang_mantis_2024, qbench,muirbench,fu2024blink}.

\vspace{5pt}
\noindent\textbf{Single-Image Evaluation.}
In Table~\ref{tab:sft-single-all}, we compare models of comparable sizes, including both closed-source~\cite{gpt4v, mm1.5} and open-source models~\cite{vila, abdin_phi3_2024, yao2024minicpmvgpt4vlevelmllm, pixtral, llama3p1,cambrian}. We report individual benchmark scores as well as the overall average score across all benchmarks, following the standard practice. 

\begin{table}[!h]
  \centering
  \resizebox{0.9\linewidth}{!}{
\begin{tabular}{@{}lcccc@{}}
\toprule
Model            & BLINK & QBench & MuirBench & Mantis-Eval \\ \midrule
\color{gray} GPT-4V~\cite{gpt4v}   & \color{gray} 51.1          & \color{gray} 73.4    & \color{gray} -  & \color{gray} 62.7\\\midrule
\textit{3B-5B Models}     &       &        &      &       \\\midrule
VILA-1.5-3B~\cite{vila}      & 39.8  & 51.7 & 29.2   & 41.9        \\
BLIP-3-4B-SI     & 42.9  & 51.9 & 35.0  & 49.3        \\
BLIP-3-4B-MI     & \textbf{47.1}  & \textbf{69.6} & \textbf{38.0}  & \textbf{56.2}        \\\midrule
\textit{10B - 15B Models} &       &    &    &             \\\midrule
VILA-1.5-13B~\cite{vila}     & 48.1  & 61.2 & 37.7  & 49.3        \\
BLIP-3-14B-SI    & 47.1  & 55.4 & 49.1  & 55.3        \\
BLIP-3-14B-MI    & \textbf{53.9}  & \textbf{73.4} & \textbf{56.2}  & \textbf{61.3}        \\ \bottomrule
\end{tabular}
\caption{\label{tab:sft-multi-all} \textbf{Evaluation on multi-image benchmarks.} VILA-1.5 models are evaluated using the same evaluation code as our models.}}
\end{table}

\vspace{5pt}
\noindent\textbf{Multi-Image Evaluation.}
In Table~\ref{tab:sft-multi-all}, we compare BLIP-3 single-image (SI) model with BLIP-3 interleaved multi-image (MI) model on multi-image benchmarks. Although the former is fine-tuned from a pre-trained model that can comprehend interleaved image-text data, it performs poorly on multi-image benchmarks, possibly because the single-image SFT causes degradation in such ability. With interleaved multi-image SFT, we see significant improvements. 


\section{Ablation Studies}
\subsection{Ablation on Stage-1 Pre-training}

\vspace{5pt}
\textbf{Few-shot Pre-training Evaluation.} To analyze the impact of pre-training configurations without the impact of fine-tuning data, we conduct ablation studies on pre-trained models with few-shot evaluation. Following~\cite{mckinzie_mm1_2024}, we randomly sample a few-shot subset from the training set and evaluate model performance. We report CIDEr scores for captioning tasks and accuracy for VQA to provide a clear comparison of few-shot performance across different pre-training configurations.

\begin{figure}[!ht]
    \centering
    \begin{subfigure}[b]{0.49\textwidth}
        \includegraphics[width=\textwidth]{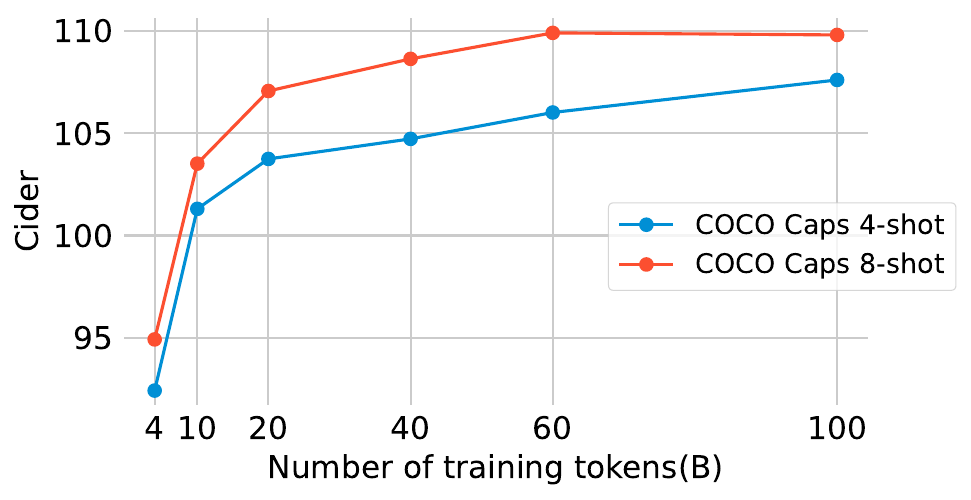}
        \caption{COCO-Caps}
        \label{fig:data_cococaps}
    \end{subfigure}
    \hfill
    \begin{subfigure}[b]{0.49\textwidth}
        \includegraphics[width=\textwidth]{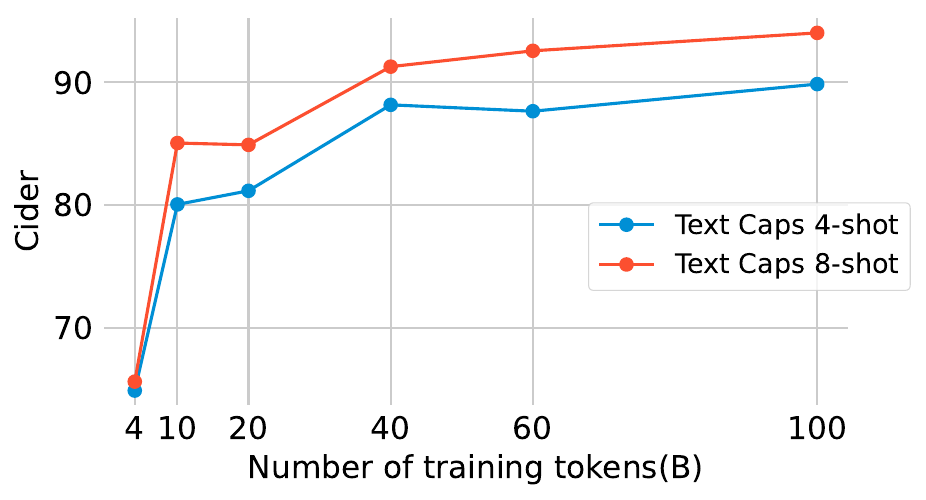}
        \caption{Text-Caps}
        \label{fig:data_textcaps}
    \end{subfigure}
    \caption{\textbf{Few-shot performance given different scales of pre-training data in Stage-1 pre-training}.}
    \label{fig:pretrain_ablation_token}
    \vspace{-3mm}
\end{figure}

\vspace{10pt}
\noindent\textbf{Scaling Pre-training Data.}
For Stage-1 pre-training, we perform an ablation study to explore the relation between the amount of pre-training data and the pre-train few-shot evaluation metrics, by varying the data scale from 2B multimodal tokens to 100B multimodal tokens. The data recipe we used here is a mixture of image caption datasets and multimodal interleaved data as detailed in \ref{fig:pretrain_datasets}. As shown in~\cref{fig:pretrain_ablation_token}, we find that scaling up the number of multimodal tokens from 2B to 60B leads to substantial gain for image-text (COCO-Caps) and OCR (Text-Caps) tasks, and further increasing the data size to 100B has moderate additional benefit in terms of few-shot evaluation metrics.

\vspace{-5pt}
\paragraph{Visual Backbones.}
We also explore if different visual backbones have an impact on the performance of vision-language tasks. We compare two types of visual encoders, DFN~\cite{dfn} and SigLIP~\cite{siglip}. As shown in \ref{tab:pretrain-ablation-visual}, we find SigLIP provides better visual representations that boost performance on OCR tasks, and we adopt SigLIP in the final model architecture as the ViT backbone. To ensure computational efficiency and a fair comparison, all models in this ablation study are pre-trained with 10B tokens.


\begin{table}[h!]
  \centering
  \resizebox{\linewidth}{!}{
    \begin{tabular}{lcccc}
    \toprule
    \textbf{Visual Backbone} & \textbf{Text-VQA} & \textbf{OK-VQA} & \textbf{COCO-Caps} & \textbf{Text-Caps} \\ 
    \midrule
    DFN     & 41.1 / 41.9  & \textbf{48.4} / \textbf{49.5}  & 107.2 / 109.4  & 78.2 / 79.9  \\
    SigLIP  & \textbf{49.1} / \textbf{50.5}  & 48.4 / 48.9  & \textbf{108.7} / \textbf{110.2}  & \textbf{84.7} / \textbf{88.6}  \\
    \bottomrule
    \end{tabular}
  }
  \caption{\textbf{Few-shot (4-shot / 8-shot) performances for different visual backbones at Stage-1 pre-training.}}
  \label{tab:pretrain-ablation-visual}
\end{table}

\subsection{Ablation on Stage-2 Pre-training}
\label{sec:stage-2-pretraining-ablation-studies}
In this section, we investigate the impact of Stage-2 pre-training. Model performance is evaluated by fine-tuning on a smaller subset ($\sim$ 1M samples) of our instruction tuning dataset for better computational efficiency. For evaluation, we focus on two representative LMM capabilities: general VQA and OCR, each of which is an averaged score over multiple relevant benchmarks.

\vspace{10pt}
\noindent\textbf{Impact of Adding Stage-2 Pre-training and with Different Resolutions.}
As shown in Table~\ref{tab:pretrain-stage2-ablation-resolution}, we use our 100B token Stage-1 pre-trained model as the baseline. We then add Stage-2 pre-training on top of it; we use half the data samples for a full Stage-2 pre-training run to maintain computational efficiency.
\looseness -1 Furthermore, we examine how varying Stage-2 pre-training resolutions (i.e., different numbers of image patches) affect performance. To ensure a fair comparison, all models in Table~\ref{tab:pretrain-stage2-ablation-resolution} are fine-tuned with a maximum of 13 image patches and follow the same fine-tuning protocol.
From the results, we observe that incorporating Stage-2 pre-training leads to a significant improvement in OCR, while general VQA performance remains stable.
\begin{table}[h!]
  \centering
  \resizebox{\linewidth}{!}{
    \begin{tabular}{cccc}
    \toprule
    \textbf{Stage-2} & \textbf{Stage-2 Maximum Resolutions} & \textbf{OCR} & \textbf{GeneralVQA} \\ 
    \midrule
    $\times$     & N/A  & 64.6 & \textbf{63.5} \\
    \checkmark & (768x768), (384x1152), (1152x384) & 65.3 & 63.2 \\
    \checkmark & (1152x1536), (1536x1152) & \textbf{67.2} & 63.3 \\
    \bottomrule
    \end{tabular}
    }
  \caption{\textbf{Stage-2 pre-training and its resolution impact.} We ablate two resolution grid settings for Stage-2 pre-training: one with up to 4 image patches, at most arranged as (2×2, 1×3, or 3×1), and another with up to 12 image patches, at most arranged as (3×4 or 4×3), with each patch sized at 384×384 pixels.}
  \label{tab:pretrain-stage2-ablation-resolution}
\end{table}

\begin{figure*}[!h]
\centering
\includegraphics[scale=0.45]{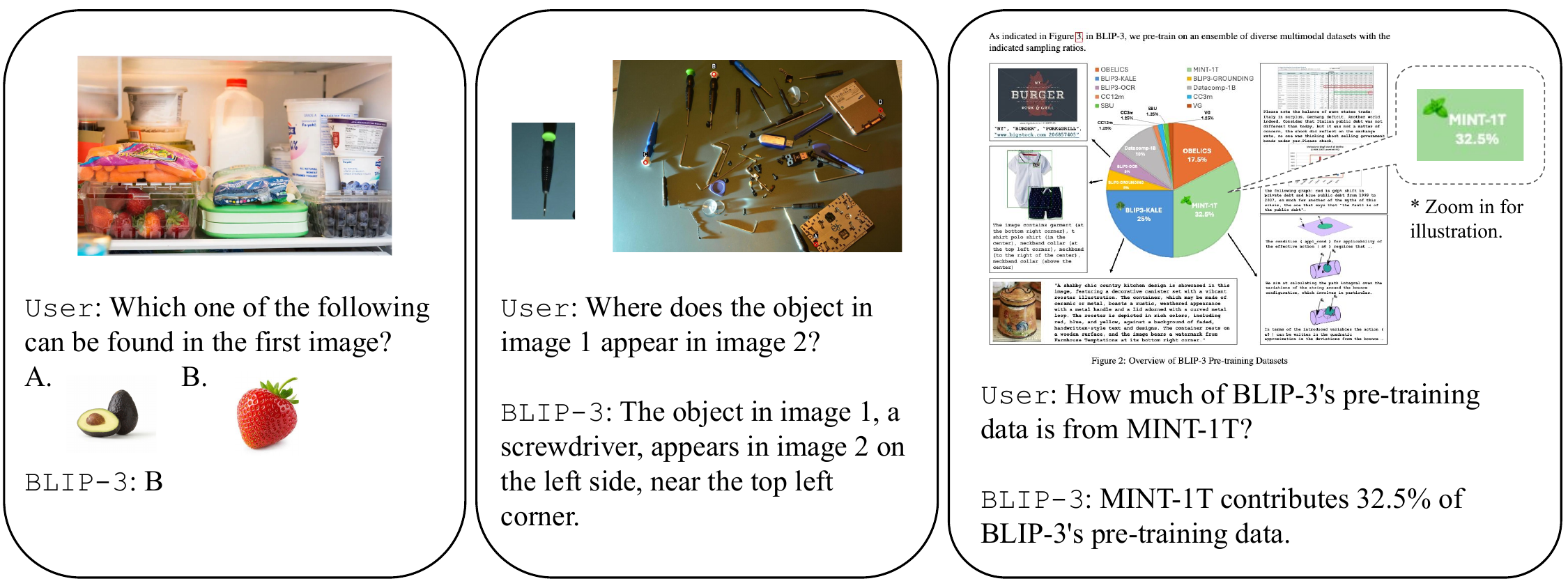}
\caption{\small{\textbf{Example model outputs of BLIP-3-4B-MI (interleaved multi-image model).} The model is capable of understanding interleaved image-text input and user queries about multiple images while maintaining the performance on single-image QAs.}}
\vspace{-10pt}
\label{fig:sft-examples}
\end{figure*}
\subsection{Ablation on Instruction Tuning}
\label{sec:sft-ablation-studies}
The ablation study in this subsection focuses on several model design choices. Evaluations are conducted on single-image fine-tuned checkpoints. Similar to section~\ref{sec:stage-2-pretraining-ablation-studies}, models in this ablation are fine-tuned on a smaller SFT subset for better computational efficiency, so the results are not directly comparable to our main results. 

\noindent\textbf{Perceiver Resampler vs. MLP.}
We experiment with two designs of vision token sampler with similar ``vision compression ratios''. Both models are trained from Stage-1 pre-training. For perceiver resample, we use 128 query tokens per image patch to achieve a compression ratio of $r_c = 729/128=5.7$ . With MLP projector, to get a similar compression ratio, we append a 2x2 max pooling layer to it to have $r_c=4$. In Table~\ref{tab:ablation-vl-projector}, we compare the two models on benchmarks focusing on various domains and report the average score on each domain. Perceiver resampler, with a slightly higher compression ratio, has a lower score on OCR, but performs better in other domains, and has a better overall score. We therefore choose to use perceiver resampler for BLIP-3, because the higher compression ratio is crucial for efficiency when handling interleaved multi-image input.

\begin{table}[!h]
\vspace{5pt}
  \centering
  \resizebox{\linewidth}{!}{
\begin{tabular}{@{}lcccc@{}}
\toprule
Vision Token Sampler & GeneralVQA & OCR  & Sci.\& Math & Hallucination \\ \midrule
Perceiver resampler  & \textbf{63.1}       & 70.4 & \textbf{43.4}        & 62.2          \\
MLP + 2x2 pooling  & 61.7       & \textbf{72.2} & 41.8        & \textbf{62.7}          \\ \bottomrule
\end{tabular}
\caption{\label{tab:ablation-vl-projector}\textbf{Ablation study on vision token sampler.} We experiment with two sampling designs with similar vision compression ratio.}}
\end{table}

\noindent\textbf{Any-Resolution Vision Token Sampling.}
Our any-resolution strategy differs from previous work~\cite{liu2024llavanext} in that every group of image embeddings (of the same image patch) is downsampled with a perceiver resampler to ensure efficiency. 
In this section, we ablate the effectiveness of our any-resolution strategy by comparing it with other designs. Models in this study are fine-tuned from the Stage-1 pre-trained base resolution models.

\begin{figure}[!h]
\centering
    \begin{subfigure}[t]{0.49\textwidth}
        \centering
        \includegraphics[scale=0.20]{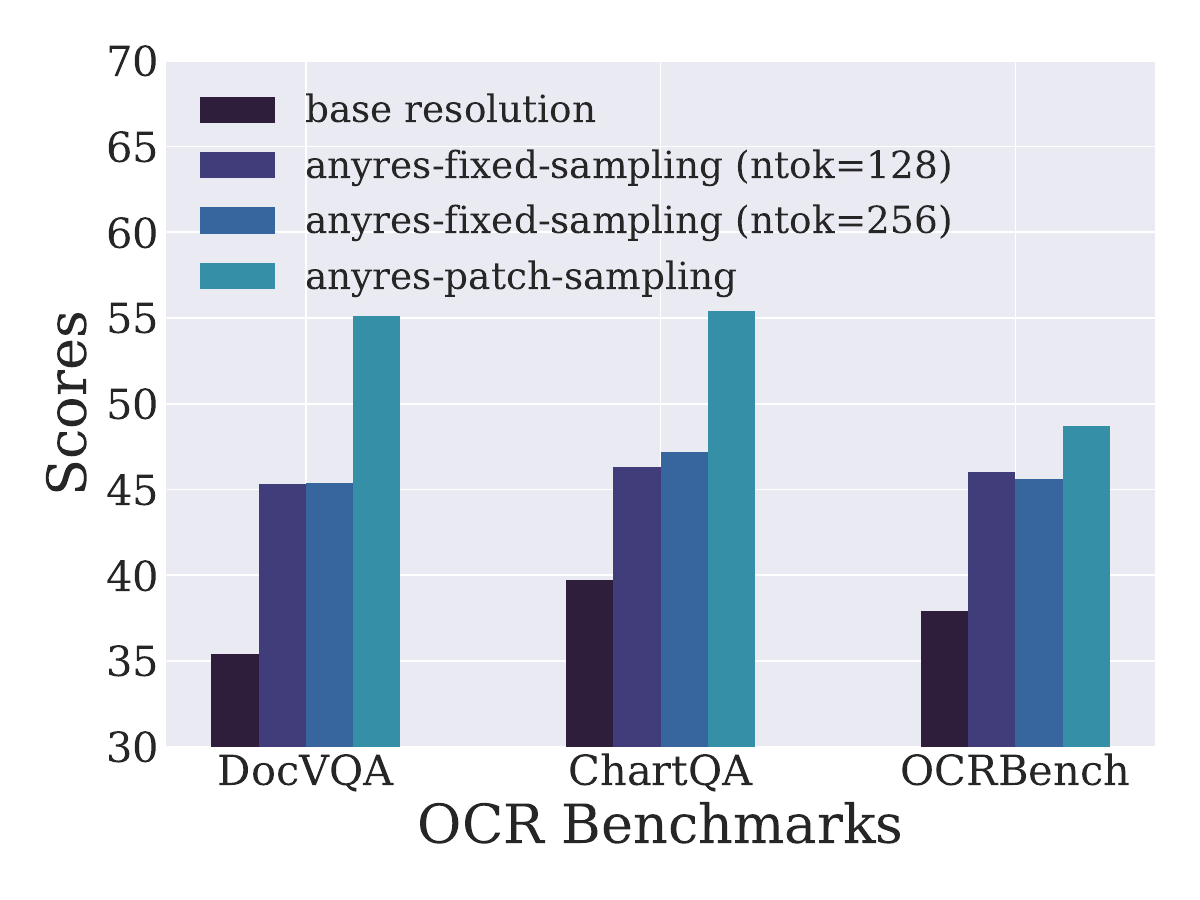}
        \caption{}
        \label{fig:sft-ablation-ocr}
    \end{subfigure}%
    \hfill
    \begin{subfigure}[t]{0.49\textwidth}
        \centering
        \includegraphics[scale=0.20]{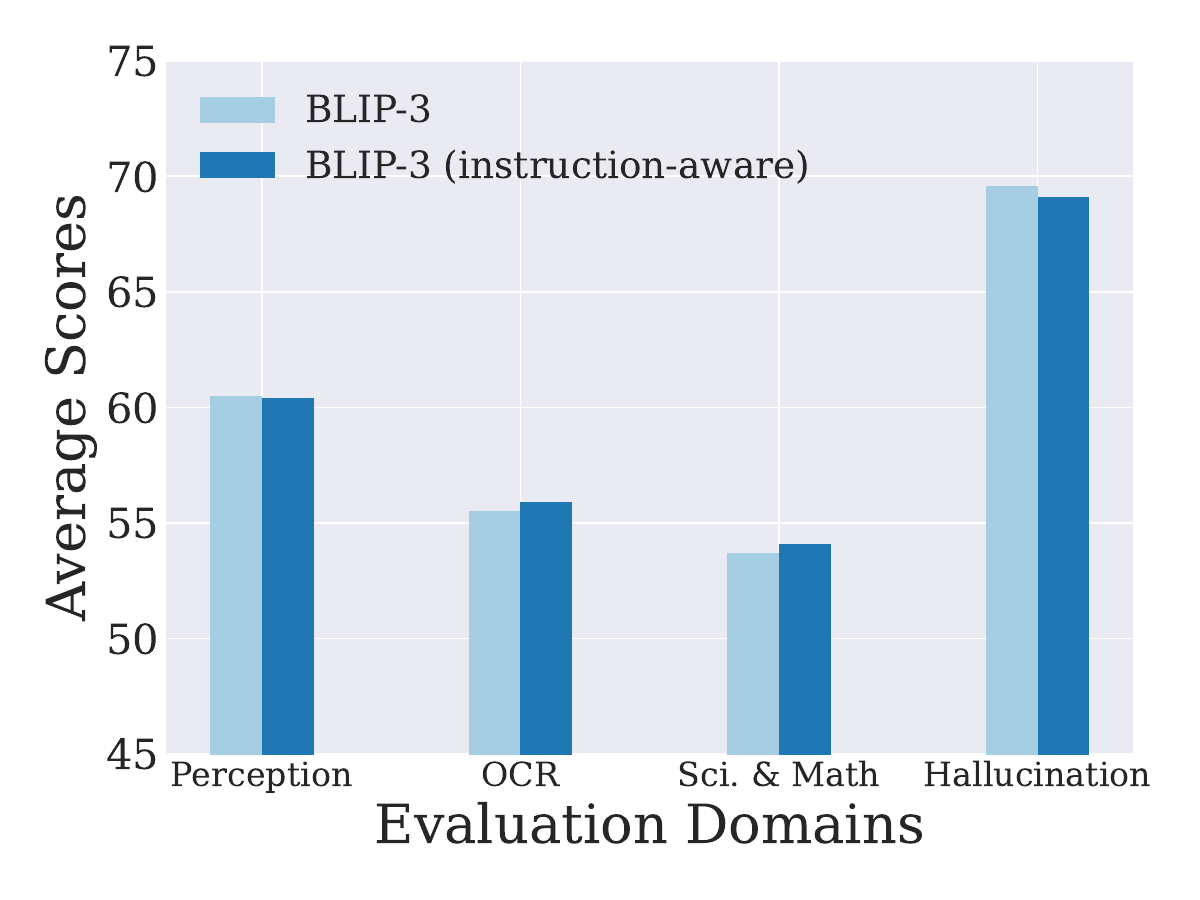}
        \caption{}
        \label{fig:sft-ablation-instruct}
    \end{subfigure}
\vspace{-5pt}
\caption{\small{\textbf{SFT ablation studies.} (a). Comparison of different vision token sampling strategies on OCR benchmarks. (b). Comparison between our model and its ``instruction-aware'' alternative. For each evaluation domain in Figure (b), we report the average score on multiple relevant benchmarks.}}
\label{fig:sft-ablation-1}
\end{figure}


\looseness -1 The ``fixed-resolution'' baseline resizes all images to the default input size of the vision encoder while keeping the original aspect ratios. Another downsampling strategy with the perceiver resampler is ``fixed sampling'' ( $\texttt{anyres-fixed-sampling}$), for which we concatenate the image embeddings from all image patches and then input them as a single sequence to the perceiver resampler to obtain the fixed number of vision tokens for the whole image.

Our evaluation of this design focuses on text-rich OCR benchmarks. From Figure~\ref{fig:sft-ablation-ocr}, we see significant improvements with the any-res image encoding strategy even with downsampled vision tokens. The fixed sampling strategy, although it shows improvements over the base resolution baseline, is not as good as the patch-wise sampling. We suspect two reasons for this: (a) With fixed sampling, the vision token compression ratio can be too high to retain text-rich visual information. (b) The perceiver resampler may not work well with a concatenation of different image embeddings.





\vspace{5pt}
\noindent\textbf{Instruction-Aware Vision Token Sampling.}
InstructBLIP~\cite{instructblip} proposes an instruction-aware Q-Former~\cite{blip2} for vision token sampling and proves its effectiveness on certain benchmarks. With the perceiver resampler, we experiment with a similar strategy by appending text instruction tokens to the query tokens of the perceiver resampler. This enables the instruction (text tokens) to interact with both query tokens and image embeddings via cross-attention. 

From the comparison in Figure~\ref{fig:sft-ablation-instruct}, we do not observe a significant difference between our model and its instruction-aware version on common task domains.
Because of the little difference we observe in this ablation study, we keep the original perceiver resampler architecture in our model for simplicity. We leave the further exploration of the ``instruction-aware'' VL connector to future works.



\section{Conclusion}
\label{sec:conclusions}
\looseness -1 We introduce BLIP-3, a comprehensive framework for training a series of open-source large multimodal models on a curated mixture of large-scale datasets. BLIP-3 LMMs achieve competitive results on a range of multimodal benchmarks. By open-sourcing BLIP-3 (4B and 14B), our curated datasets, and our training code, we hope to empower the research community with better accessible multimodal foundation models and datasets, allowing practitioners to explore further and advance the potential and emergent abilities of LMMs.

{
    \small
    \bibliographystyle{ieeenat_fullname}
    \bibliography{main}
}

\end{document}